\documentclass[sigconf]{acmart}

\usepackage{graphicx}
\usepackage{amsmath}
\usepackage{amssymb}
\usepackage{booktabs}
\usepackage{adjustbox}
\usepackage{makecell}
\usepackage{indentfirst}

\copyrightyear{2022}
\acmYear{2022}
\setcopyright{acmcopyright}
\acmConference[MM '22]{Proceedings of the 30th ACM International Conference on Multimedia}{October 10--14, 2022}{Lisboa, Portugal}
\acmBooktitle{Proceedings of the 30th ACM International Conference on Multimedia (MM '22), October 10--14, 2022, Lisboa, Portugal}
\acmPrice{15.00}
\acmDOI{10.1145/3503161.3549201}
\acmISBN{978-1-4503-9203-7/22/10}
\usepackage[capitalize]{cleveref}
\crefname{section}{Sec.}{Secs.}
\Crefname{section}{Section}{Sections}
\Crefname{table}{Table}{Tables}
\crefname{table}{Tab.}{Tabs.}

\AtBeginDocument{%
  \providecommand\BibTeX{{%
    \normalfont B\kern-0.5em{\scshape i\kern-0.25em b}\kern-0.8em\TeX}}}

\acmSubmissionID{2}

\begin{document}

\title{Can Language Understand Depth?}


\author{Renrui Zhang}
\authornote{Both authors contributed equally to this research.}
\email{1700012927@pku.edu.cn}
\affiliation{%
  \institution{Peking University}
  \country{China}
}
\author{Ziyao Zeng}
\email{zengzy@shanghaitech.edu.cn}
\authornotemark[1]
\affiliation{%
  \institution{ShanghaiTech University}
  \country{China}
}

\author{Ziyu Guo}
\email{2101210573@pku.edu.cn}
\affiliation{%
  \institution{Peking University}
  \country{China}
}

\author{Yafeng Li}
\email{liyafeng770729@126.com}
\affiliation{%
  \institution{Baoji University of Arts and Science}
  \country{China}
}



\begin{abstract}
Besides image classification, Contrastive Language-Image Pre-training (CLIP) has accomplished extraordinary success for a wide range of vision tasks, including object-level and 3D space understanding. However, it's still challenging to transfer semantic knowledge learned from CLIP into more intricate tasks of quantified targets, such as depth estimation with geometric information. In this paper, we propose to apply CLIP for zero-shot monocular depth estimation, named \textbf{DepthCLIP}. We found that the patches of input image could respond to a certain semantic distance token and then be projected to a quantified depth bin for coarse estimation. Without any training, our DepthCLIP surpasses existing unsupervised methods and even approaches the early fully-supervised networks. To our best knowledge, we are the first to conduct zero-shot adaptation from the semantic language knowledge to quantified downstream tasks and perform zero-shot monocular depth estimation. We hope our work could cast a light on the future research. The code is available at \url{https://github.com/Adonis-galaxy/DepthCLIP}. 

\end{abstract}



\maketitle
\section{Introduction}




\label{sec:intro}

 Multi-modality learning has long been a fundamental problem, for which a lot of proposed models utilized language knowledge to assist vision tasks and showed inspiring outcomes. Therein, Contrastive Language-Image Pre-Training (CLIP)~\cite{clip} exerted powerful transfer ability and achieved promising performance on zero/few-shot image classification~\cite{gao2021clip_cls,zhang2021vt_cls,zhang2021tip_cls,zhou2021learning_cls}, object detection~\cite{zhou2022detecting_detect,rao2021denseclip_detect_seg}, semantic segmentation~\cite{zhou2021denseclip_seg,rao2021denseclip_detect_seg} and others~\cite{zhang2022pointclip}. Further, for those vision tasks, CLIP is only required to recognize visual signals from a high level, and how to apply its pre-trained semantic language knowledge to quantified vision tasks, e.g., depth estimation, has not been further explored.

Monocular depth estimation is a vital task in the industrial field, which serves as an essential component in various tasks like monocular 3D object detection~\cite{liu2020smoke,wang2021fcos3d,zhang2022monodetr} or point cloud reconstruction from images~\cite{zeng2018inferring}. It normally relies on dense depth labels to train a network that extracts semantic relations within an image and regresses pixel-wise depth value. However, training networks from scratch supervised by dense labels severely hinders the efficient deployment for application. Also, it is quite costly to collect and annotate the large-scale datasets, such as NYUV2~\cite{silberman2012indoor}. Some of the unsupervised methods need extra data like single-view video~\cite{zhang2020unsupervised,mahjourian2018unsupervised} to capture time-consistent constraint, or demand 3D priors~\cite{jiang2021plnet} for better spatial modeling. We then ask the question: can we avoid the cost for both training models and collecting data by using semantic language knowledge learned by CLIP~\cite{clip}?

\begin{figure}[t]
  \centering
  \vspace{0.3cm}
\includegraphics[width=0.48\textwidth]{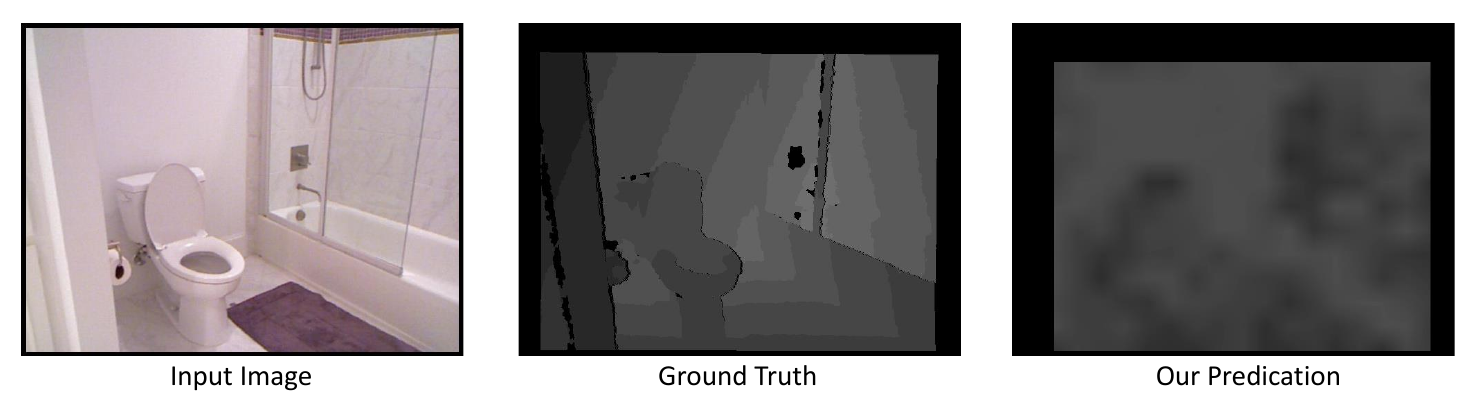}
  \caption{Visualizations of our DepthCLIP depth estimation. We require no extra training and directly transfer CLIP~\cite{clip} for zero-shot prediction.}
    \label{fig:domain_gap}
    \vspace{-0.5cm}
\end{figure}

For the first time, we explore the relationship between language and depth and apply CLIP~\cite{clip} to monocular depth estimation. CLIP~\cite{clip} trains an visual encoder and a textual encoder jointly by contrastive loss in the embedding space between both modalities, which narrows the cosine distance of paired image and text features while pushing away the others. The current transfer mode of CLIP only utilizes its pre-trained classification ability, such as classifying each localized object for detection and categorizing each pixel for segmentation. In contrast, during contrastive pre-training, CLIP is not likely to learn knowledge for regressing quantitive values. Thus, it's impossible to directly output depth prediction from the CLIP model for depth estimation. Considering this, we convert the depth value regression to a distance classification task that the CLIP is only required to understand "the object is close/far.", but not "the object is 5 meters away."

In our experiments, we show that the pre-trained CLIP model is able to make responses to the pre-defined distance concepts (like close or remote) for different patches of an image. Then, we map each distance concept to a certain quantitive depth bin and linearly combine the multi-bin depth values weighted by language-patch similarities to produce the depth estimation for a patch. For simplicity, pixels within the same patch would be given the same predicted depth value. By this, our DepthCLIP could conduct zero-shot depth estimation using pre-trained CLIP without any fine-tuning or extra training data, which exceeds existing unsupervised transferred methods and only have a light gap with early fully-supervised models. For analysis, we visualize the distance response of image patches to show that DepthCLIP could coarsely capture the semantic distance concepts. Besides, we test class-dependent bins to verify that by setting such class-class-dependent bins, DepthCLIP could achieve further improvement for wider practical applications. We summarize our contributions as below:

\begin{itemize}
    \item We propose DepthCLIP, which is the first to perform zero-shot adaptation from semantic language knowledge to monocular depth estimation task. 
    
    \item We build the bridge between CLIP's semantic knowledge and quantified depth value prediction, which casts a light on future research.
    
    \item To illustrate our effectiveness, we experiment DepthCLIP on NYU Depth v2~\cite{silberman2012indoor}, which exceeds existing unsupervised methods and draws near to some fully-supervised models.
\end{itemize}

\section{Related Work}
\label{sec:related work}

\paragraph{\textbf{Vision-Language Models}}

Vision-Language learning has been widely studied and achieved remarkable outcomes to benefit visual representation learning, especially for zero-shot domain transfer in a wide range of downstream tasks. 
Contemporarily, driven by extensive image-text pairs emerging on the Internet, CLIP~\cite{clip} and ALIGN~\cite{align} have raised a revolution in vision language learning, in which the former uses 400 million noisy image-text pairs and the latter uses 1.8 billion noisy image-text pairs. 

\subparagraph{\textbf{CLIP}}~\cite{clip} was originally designed for zero-shot image classification. During training, in each batch, CLIP encodes images and corresponding texts into feature space, then maximizes the similarity between corresponding pairs, while minimizing the rest. To conduct zero-shot classification during testing, prompts like “A photo of [class]” would be formed, and the class that has max similarity between prompt features and image features will serve as the prediction. After its remarkable success in zero-shot image classification, it has been applied to few-shot image classification by ~\cite{gao2021clip_cls,zhang2021vt_cls,zhang2021tip_cls,zhou2021learning_cls}, to image semantcation by ~\cite{zhou2021denseclip_seg,rao2021denseclip_detect_seg}, and to object detection by ~\cite{zhou2022detecting_detect,rao2021denseclip_detect_seg}.
However, solving those tasks requires only the recognition ability. Image classification requires learning generic representation for a single image. Segmentation requires distinguishing between various pixels belonging to different classes(semantic segmentation), objects(instance segmentation), or both(panoptic segmentation). As for object detection, CLIP is only responsible for object-level classification rather than localization. Therefore, no work has been done for CLIP to solve quantified tasks predicting continuous values.

\vspace{0.5cm}
\begin{figure*}[t!]
  \centering
    \includegraphics[width=0.9\textwidth]{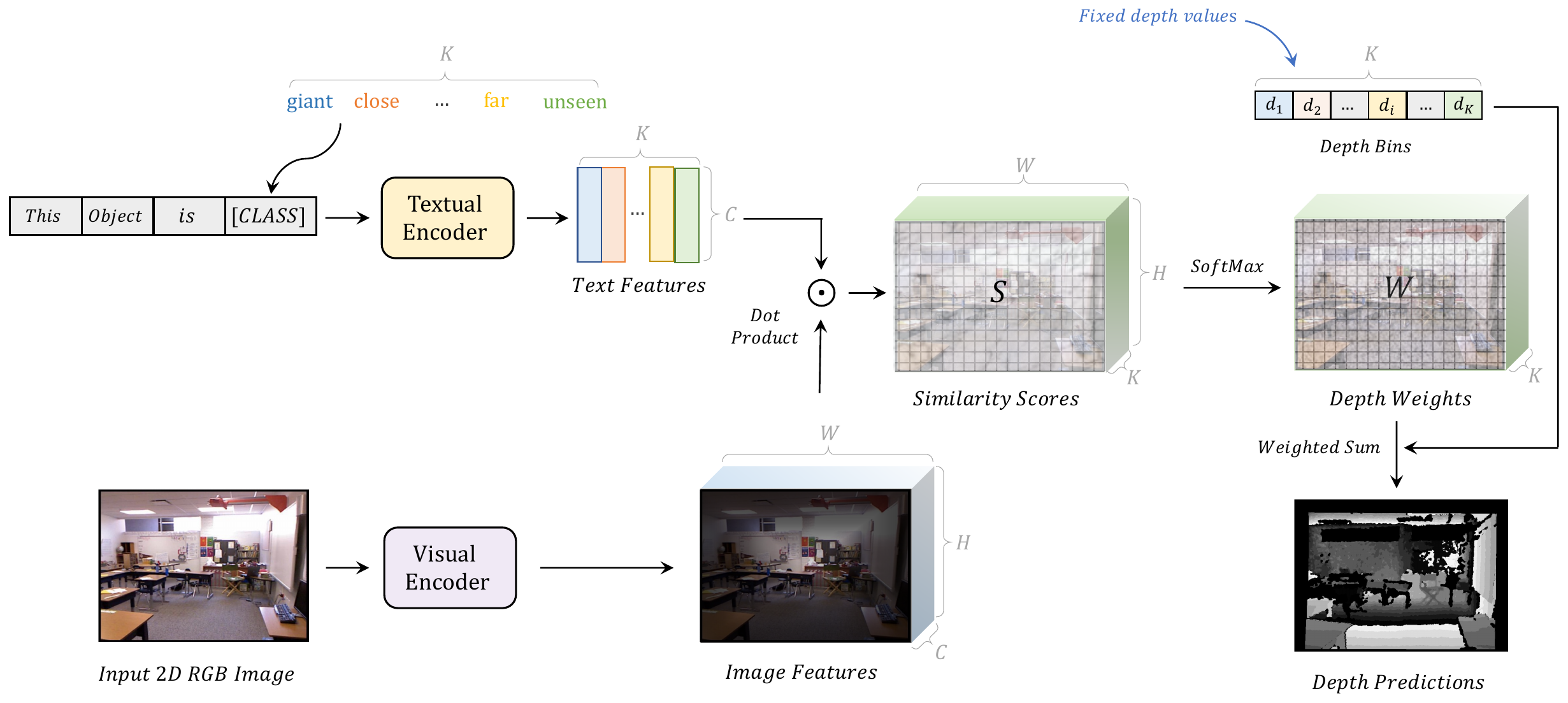}   \caption{Pipeline of DepthCLIP.}
    \label{fig:model}
\end{figure*}

\paragraph{\textbf{Monocular Depth Estimation}}
Monocular depth estimation aims to predict pixel-wise depth value from a single input image. It is a challenging task due to the lack of stereo information that is the key to solve geometry constrain for depth estimation. As a result, the model could only utilize semantic relationships extracted from the input image and semantic prior leaned from training data to solve the problem. 

\subparagraph{\textbf{Fully-supervised Methods}} could learn from ground-truth depth maps during training to master semantic prior and extract semantic relationship. 
DORN~\cite{dorn} proposes a deep convolutional neural network with a spacing-increasing depth discretization strategy and recasts depth learning as an ordinal regression problem.
RPSF~\cite{mel2022end} introduces a differentiable physical model of the aperture mask and simulates the camera imaging pipeline preciously. Thanks to the recent booming of transformer in computer vision community~\cite{tran_1,tran_2,tran_3,tran_4}, some methods also adopt such architectures for monocular depth estimation. For example, ASTransformer~\cite{ast} designs an Attention-based Up-sample Block to recompense the detailed texture features. DepthFormer~\cite{li2022depthformer} proposes a hierarchical aggregation and heterogeneous interaction modules to build up affinities between features and models. To sum up, all the fully-supervised methods have achieved astonishing performance on monocular depth estimation because of their dedicated designed structures and capacity to model semantics prior during training. However, gathering fully-annotated data would be costly and labor-consuming, which constrains its scalability.

\subparagraph{\textbf{Unsupervised Methods}} construct pre-text tasks delicately to teach models to discover the semantic affinity within a monocular image. Considering that videos could provide temporal constraints for geometry,~\cite{zhang2020unsupervised,mahjourian2018unsupervised} learn depth knowledge from ego-motion of unlabelled monocular videos. Besides, ~\cite{left-right} utilizes widely available binocular stereo images for self-training, which exploits epipolar geometry constraints to generate disparity
images for depth predication. Moreover, without using additional data from other modalities (video or stereo images), ~\cite{jiang2021plnet} introduces plane and line priors
to enhance the unsupervised monocular depth estimation. Overall, all previous methods need additional vision-based data or geometry prior to help with depth estimation. Our work is the first to use language semantic information leaned by CLIP~\cite{clip} to conduct monocular depth estimation, without any explicit geometry priors.

\section{Method}
\label{sec:method}

As shown in Figure\ref{fig:model}, our model utilizes the pre-trained knowledge from CLIP to project the semantic response of each patch into a certain depth bin, and linearly combine those depth values to obtain the final predication. We introduce and discuss three parts respectively below: Image Encoding, Text Encoding, and Depth Projection and Combination.

\subsection{Image Encoding}
Given a monocular RGB image $I$, we feed it into the visual encoder without the final pooling layer. Then we acquire its last-layer $C$-dimensional feature map $\mathbf{F}_{img}$, formulated as, 
\begin{equation} 
\mathbf{F}_{img} = \mathrm{VisualEncoder}(I)\ \in \mathbb{R}^{HW\times C}
\end{equation} 
Each spot of $\mathbf{F}_{img}$ would be given a depth forecast based on its reaction toward semantic language tokens. Since the pre-trained visual encoder of CLIP~\cite{clip} is received by molding a classification pre-text task, each site of the feature map before pooling would grasp regional semantic information, and pooling would assemble the local knowledge to develop a global interpretation of the given image. As a result, each location of the feature map would preserve its surrounding semantic details, and be responded to by text token to obtain its depth approximation.

\subsection{Text Encoding}
Due to the contrastive pre-text task of CLIP, the textual encoder of CLIP~\cite{clip} would project similar semantic tokens to the neighborhood of image features. In DepthCLIP, we utilize prompts like 'This object is {distance class}' and apply the semantic tokens {close, far, remote} that substitute distance class to form $K$ semantic tokens. Those tokens $T$ will pass through the textual encoder of CLIP~\cite{clip} into latent space by \begin{equation} 
\mathbf{F}_{text} = \mathrm{TextualEncoder(T)}\ \in \mathbb{R}^{K\times C},
\end{equation}
 where the dimension $C$ equals $\mathbf{F}_{img}$. We then calculate the cosine similarity between semantic tokens' features $\mathbf{F}_{text}$ and image features $\mathbf{F}_{img}$ to acquire similarity scores by
\begin{equation} \mathbf{S}=\frac{\mathbf{F}_{text} \cdot \mathbf{F}_{img}^\top}{\|\mathbf{F}_{text}\|\|\mathbf{F}_{img}\|}\ \in \mathbb{R}^{K\times HW}
\end{equation}

\subsection{Depth Projection and Combination}
After obtaining similarity scores $\mathbf{S}$ ($dim=H*W*K$), they ought to be cast to quantified depth to receive depth prediction. That is to say, "close" shall be projected to "1m". We apply a temperature softmax function among such similarity scores $S$ to obtained linear combination weight by: 
\begin{equation}
\mathbf{W}=\frac{e^{\mathbf{S}_{:,:,i}}}{\sum_{j=1}^{K} e^{\mathbf{S}_{:,:,j}}}, \quad \text { for } i=1, \ldots, K
\end{equation}
Then, we employ such weights to linearly combined depth bins $\mathbf{d}$ ($dim=K*1$, like [$d_1$=1.00m, $d_2$=2.00m, $d_3$=3.00m, ...]) to obtain the final depth prediction $\mathbf{D_{pred}}$ ($dim=H*W*1$) for each location of the image feature map, which is aligned with patches in the image:
\begin{equation}
\mathbf{D_{pred}}=\sum_{i=1}^{K} \mathbf{W}_{:,:,i} * \mathbf{d}_i.
\end{equation}
Where $\mathbf{D_{pred}}_{i,j}$ is the depth of the patch in $i$ th  row and $j$ th column. All pixels belonging to such certain patch will be presented with the same depth prediction $\mathbf{D_{pred}}_{i,j}$, where $i \in [1,H], j \in [1,W]$. 

\section{Experiments}
\label{sec:experiments}
To demonstrate the effectiveness of our zero-shot training-free depth estimator, we conduct ample experiments examining different facets of our approach. After introducing the implementation details, we evaluate our methods on a mainstream challenging benchmark——NYU Depth v2~\cite{silberman2012indoor}. Ablation studies and visualization are provided to offer a more thorough investigation of our method.

\subsection{Datasets}
We evaluate our method on NYU Depth v2~\cite{silberman2012indoor}. The dataset consists of 120K pairs of RGB and depth images, in which all image pairs are taken by the Microsoft Kinect sensor under 464 indoor scenes with a resolution of 480×640. Depth range for each pixel is 0-10m. The same training/testing split configuration are applied following ~\cite{lee2019big,eigen2014depth}. The training set covers 36,253 images from 249 scenes, while the testing set contains 654 images from the remaining 215 scenes. To remove frames, all samples are 
cropped to the resolution of 416×512 with the identical configuration in ~\cite{song2021monocular}.

\subsection{Implementation Details}
We implement our model with the PyTorch framework. Our image and textual encoders employ the pre-trained ResNet-50~\cite{resnet} of CLIP~\cite{clip}. For semantic prompts, we tested various kinds of hand-craft prompts, and pick "This object is [distance class]". For semantic distance classes, we investigated diverse combinations and select  ['giant', 'extremely close', 'close','not in distance','a little remote', 'far','unseen'], 7 semantic bins in total. Each of which aligns with a depth bin of [1.00, 1.50, 2.00, 2.25, 2.50, 2.75, 3.00]. We set this setting for our main experiments since the range of indoor depth could be properly captured under such proper numbers of semantic and depth bins. The temperature of the final softmax function is set to 0.1.

\subsection{Evaluation Metrics}
The evaluation metrics are following previous works ~\cite{ast}. We compare our method quantitatively with metrics listed below: mean absolute relative error (rel), root mean square error (rmse), absolute error in log space $\left(\log _{10}\right)$, and threshold accuracy $\left(\delta_{i}\right)$. The formula of each metric has been itemized below:

\begin{equation}
\begin{aligned}
&\text { rel }=\frac{1}{n} \sum_{p}^{n} \frac{\left|y_{p}-\hat{y}_{p}\right|}{\hat{y}_{p}}, \quad \text { rmse }=\sqrt{\left.\frac{1}{n} \sum_{p}^{n}\left(y_{p}-\hat{y}_{p}\right)^{2}\right)} \\
&\log _{10}=\frac{1}{n} \sum_{p}^{n}\left|\log _{10}\left(y_{p}\right)-\log _{10}\left(\hat{y}_{p}\right)\right| \\
&\delta=\% \text { of } y_{p} \text { s.t. } \max \left(\frac{y_{p}}{\hat{y}_{p}}, \frac{\hat{y}_{p}}{y_{p}}\right)=\delta<t h r \text { for } t h r=1.25,1.25^{2}, 1.25^{3}
\end{aligned}
\end{equation}

\subsection{Quantified Results}
Table \ref{tab:nyu_results} shows our results compared with other monocular depth estimation methods. The table is divided by different supervisions and pre-training datasets. The lower bound is obtained by randomly making predication for each pixel within the depth range 0-10m. It could be noticed that despite the gap between our method with fully-supervised methods, DepthCLIP exceeds the lower bound by a large margin, even surpassing other zero-shot transferring methods that are pre-trained on the dataset especially prepared for monocular depth estimation(unsupervised KITTI monocular video~\cite{geiger2013vision_kitti}). Besides, the performance of our DepthCLIP has been highly close to some fully-supervised methods like Make3D~\cite{make3d}, even exceeding it in terms of rmse(1.186 of ours compared with 1.214 of Make3D~\cite{make3d}).

\begin{table*}[t]
\vspace{0.3cm}
\centering
\begin{adjustbox}{width=0.9\linewidth}
	\begin{tabular}{l|cc|ccc|ccc}
	\toprule
		Method & Pre-training & Supervision & $\delta<1.25 \uparrow$ & $\delta<1.25^{2} \uparrow$ & $\delta<1.25^{3} \uparrow$ & rel $\downarrow$ & $\log _{10} \downarrow$ & rmse $\downarrow$ \\ \midrule
		
		 Make3D~\cite{make3d} &- &depth  &0.447 &0.745 &0.897 &0.349 & - &1.214\\
		 
		 DORN~\cite{dorn}&- &depth &0.828 &0.965 &0.992 &0.115 &0.051 &0.509\\
		 
		 ASTransformer~\cite{ast} &- &depth &0.902 &0.985 &0.997,  &0.103 &0.044 &0.374\\
		 
		 DepthFormer~\cite{li2022depthformer}  &- &depth &0.921	&0.989	&\textbf{0.998} &0.096	&0.041 &0.339\\

		 RPSF~\cite{mel2022end}  &- &depth &\textbf{0.952} &\textbf{0.989} &0.997	&\textbf{0.072}	&\textbf{0.029}		&\textbf{0.267} \\

		 \midrule
		 Lower Bound &- &- &0.140 &0.297 &0.471 &1.327 &0.323 &2.934\\
		 
		 vid2depth~\cite{mahjourian2018unsupervised}  &KITTI video~\cite{geiger2013vision_kitti} &0-shot &0.268	&0.507	&0.695 &0.572 &- &1.637\\
		 
 		 Zhang et al.~\cite{zhang2020unsupervised}  &KITTI video~\cite{geiger2013vision_kitti} &0-shot &0.350	&0.617	&0.799 	&0.513 &0.529 &1.457\\

	    \textbf{DepthCLIP} 	&CLIP~\cite{clip} &0-shot &\textbf{0.394} &\textbf{0.683} &\textbf{0.851} &\textbf{0.388} &\textbf{0.156} &\textbf{1.167} \\
	    
	\bottomrule
	\end{tabular}
\end{adjustbox}
\vspace{0.3cm}
\caption{Performance of Monocular Depth Estimation on NYU Depth v2~\cite{silberman2012indoor}. The table is divided by different supervisions and pre-training datasets. Lower bound is obtained by randomly making predication for each pixel within depth range 0-10m.}
\vspace*{0pt}
\label{tab:nyu_results}
\end{table*}

\subsection{Class-dependent Depth Bin}
Our DepthCLIP requires attaching a quantified depth bin to each semantic language token, then linearly combined to obtain final predication. Shown as Table \ref{fig:domain_gap}, images from different classes possess different depth distribution. CLIP~\cite{clip} can capture semantic distance relationships within one image, but patches holding different distances of various scenes could be mapped to the same semantic concept. In other words, the same semantic token should be mapped to different class-dependent quantified depth bins in different scenes, to achieve better performance. Due to the limit of time, we remain the same bin partition for all classes, and conduct ablation to examine the effectiveness of using class-dependent bin partitioning.

Shown as Table \ref{tab:class_bin_ablation}, we select the bathroom class, the classroom class, and all classes of NYU Depth v2~\cite{silberman2012indoor} to test class-dependent depth bin partitioning. Each depth bin partition is also tested with all classes of NYU Depth v2~\cite{silberman2012indoor} to serve as a comparison. We could notice that by depth bin partition based on scene class of test image, compared with evaluating with original shared by all classes for NYU Depth v2~\cite{silberman2012indoor} used in Table \ref{tab:nyu_results}, we could achieve a remarkable performance gain. On the other hand, different classes have different best bin partitions. That is to say, we could learn a class-dependent bin partition, and conduct zero-shot classification using CLIP~\cite{clip} to match images with corresponding depth bin partitions to achieve better results, which would be a worthy future research direction.

\begin{figure}[t]
  \centering
\includegraphics[width=0.4\textwidth]{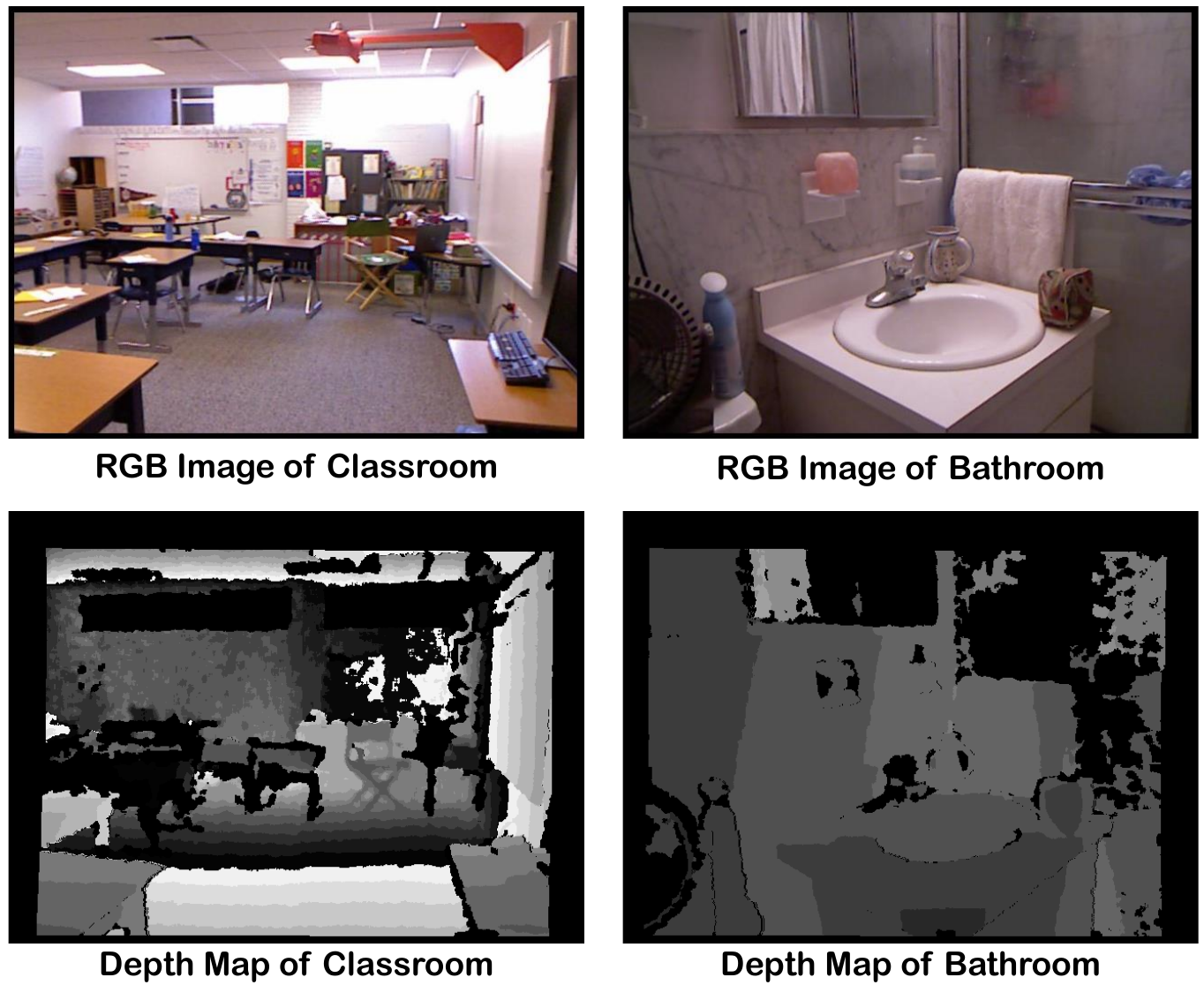}
  \caption{Depth distribution gap between different classes.}
    \label{fig:domain_gap}
\end{figure}

\begin{table}[t]
\centering
\begin{adjustbox}{width=0.7\linewidth}
	\begin{tabular}{l|c}
	\toprule
		Bin partition & Depth bin partition details (in meters)\\ \midrule
		
		 Original bin &[1.00, 1.50, 2.00, 2.25, 2.50, 2.75, 3.00]\\
		 
		 Class-dependent 1 &[1.00, 2.00, 2.25, 2.50, 2.75, 3.00, 4.00]\\
		 
		 Class-dependent 2 &[1.00, 1.50, 2.00, 2.50, 3.00, 3.50, 4.00]\\
		 
		 Class-dependent 3 &[1.00, 1.25, 1.50, 1.75, 2.00, 2.25, 2.50]\\
		 
		 Class-dependent 4 &[2.00, 2.50, 3.00, 3.25, 3.50, 3.75, 4.00]\\

	\bottomrule
	\specialrule{0em}{7pt}{7pt}
	\end{tabular}

\end{adjustbox}

\begin{adjustbox}{width=\linewidth}
	\begin{tabular}{l|ccc|ccc}
	\toprule
		Class: Bathroom & $\delta<1.25 \uparrow$ & $\delta<1.25^{2} \uparrow$ & $\delta<1.25^{3} \uparrow$ & rel $\downarrow$ & $\log _{10} \downarrow$ & rmse $\downarrow$ \\ \midrule
		
		 Original bin &0.333 &0.631 &0.814 &0.549 &0.175 &0.922\\
		 
		 Class-dependent 1 &0.248 &0.490 &0.699 &0.754 &0.219 &1.237\\
		 
		 Class-dependent 2 &0.236 &0.460 &0.675 &0.801 &0.229 &1.308\\
		 
		 \textbf{Class-dependent 3} &\textbf{0.425} &\textbf{0.723} &\textbf{0.893} &\textbf{0.373} &\textbf{0.141} &\textbf{0.745}\\
		 
		 Class-dependent 4 &0.129 &0.302 &0.535 &1.072 &0.287 &1.682\\
		 \midrule
		 
		 Best partition's gain &+0.092 &+0.092 &+0.079 &-0.176 &-0.034 &-0.177\\

	\bottomrule
	\specialrule{0em}{7pt}{7pt}
	\end{tabular}
\end{adjustbox}

\begin{adjustbox}{width=\linewidth}
	\begin{tabular}{l|ccc|ccc}
	\toprule
		Class: Classroom & $\delta<1.25 \uparrow$ & $\delta<1.25^{2} \uparrow$ & $\delta<1.25^{3} \uparrow$ & rel $\downarrow$ & $\log _{10} \downarrow$ & rmse $\downarrow$ \\ \midrule
		
		 Original bin  &0.308 &0.533 &0.742 &\textbf{0.372} &0.193 &1.826\\
		 
		 Class-dependent 1  &\textbf{0.312} &0.565 &0.820 &0.383 &0.179 &1.694\\
		 
		 Class-dependent 2  &0.310 &0.583 &0.830 &0.397 &0.175 &1.636\\
		 
		 Class-dependent 3  &0.231 &0.452 &0.600 &0.407 &0.246 &2.138\\
		 
		 \textbf{Class-dependent 4}  &0.276 &\textbf{0.637} &\textbf{0.844} &0.461 &\textbf{0.173} &\textbf{1.544}\\
		 
		 \midrule
		 
		 Best partition's gain &-0.032 &+0.104 &+0.102 &+0.088 &-0.020 &-0.282\\
		 
	\bottomrule
	\specialrule{0em}{7pt}{7pt}
	\end{tabular}
\end{adjustbox}

\begin{adjustbox}{width=\linewidth}
	\begin{tabular}{l|ccc|ccc}
	\toprule
	Class: All & $\delta<1.25 \uparrow$ & $\delta<1.25^{2} \uparrow$ & $\delta<1.25^{3} \uparrow$ & rel $\downarrow$ & $\log _{10} \downarrow$ & rmse $\downarrow$ \\ \midrule
		
		 \textbf{Original bin}  &\textbf{0.394} &\textbf{0.683} &\textbf{0.851} &0.388 &\textbf{0.156} &\textbf{1.167} \\
		 
		 Class-dependent 1  &0.373  &0.653  &0.828  &0.467  &0.166  &1.228\\
		 
		 Class-dependent 2  &0.366  &0.641  &0.819  &0.496  &0.170  &1.248\\
		 
		 Class-dependent 3  &0.333  &0.621  &0.818  &\textbf{0.353}  &0.176  &1.290\\
		 
		 Class-dependent 4  &0.288  &0.548  &0.752  &0.663  &0.201  &1.439\\
		 
		 \midrule
		 
		 Best partition's gain &- &- &- &- &- &-\\
		 
	\bottomrule
	\specialrule{0em}{7pt}{7pt}
	\end{tabular}
\end{adjustbox}
\caption{Ablations of class-dependent depth bin. Perform in bathroom, classroom, and all classes of NYU Depth v2~\cite{silberman2012indoor}. The uppermost table shows partition details of each depth bin partition, the lower three tables show the performance of each partition in the bathroom, classroom, and all classes. The original bin is the one used for Table \ref{tab:nyu_results} to evaluate the entire dataset among all classes. Performance gain between the original bin and best class-dependent bin is listed in the last row, while the best class-dependent bin is bold.}
\label{tab:class_bin_ablation}
\vspace*{-0.5cm}
\end{table}

\subsection{Prompt Design}
We evaluate our method under different prompt designs on all classes of NYU Depth v2~\cite{silberman2012indoor}. Shown as Table \ref{tab:prompt_design}, our method achieves the best performance under the original prompt design, which is used for the results we report for Table \ref{tab:nyu_results}. It is worth noticing that, the performance gap among different prompt settings does not vary too much, especially not as sensitive as the depth bin partition shown in Table \ref{tab:class_bin_ablation}. This outcome is caused by similar semantic meanings inhabiting within different semantic tokens, like "giant" and "extremely close" express a similar meaning, and a patch containing a close object would be projected to both tokens, with only response intensity varying a little for different prompt designs. In this way, the major part of the semantic response score will remain stable across different hand-craft prompt designs. It reveals that our method is robust to hand-craft prompt design, which means it would be time-saving to tune a suitable prompt for our method.

\begin{table*}[t]
\centering
\begin{adjustbox}{width=0.65\linewidth}
	\begin{tabular}{l|c}
	\toprule
		Prompt number & Prompt design details (in semantic token words)\\ \midrule
		
		 Original prompt & [giant, extremely close, close, not in distance, a little remote, far, unseen] \\
		 
		 Prompt 1 &[extremely close, close, middle, a little far, far, quite far, unseen] \\
		 
		 Prompt 2 & [extremely close, very close, close, a little close, a little far, far, unseen] \\
		 
		 Prompt 3 &[giant, close, a little close, not in distance, a bit remote, far, unseen] \\

	\bottomrule
	\specialrule{0em}{7pt}{7pt}
	\end{tabular}
\end{adjustbox}

\begin{adjustbox}{width=0.65\linewidth}
	\begin{tabular}{l|ccc|ccc}
	\toprule
		Prompt number & $\delta<1.25 \uparrow$ & $\delta<1.25^{2} \uparrow$ & $\delta<1.25^{3} \uparrow$ & rel $\downarrow$ & $\log _{10} \downarrow$ & rmse $\downarrow$ \\ \midrule
		
		 \textbf{Original prompt} &\textbf{0.394} &\textbf{0.683} &\textbf{0.851} &0.388 &\textbf{0.156} &\textbf{1.167}\\
		 
		 Prompt 1 &0.341  &0.623  &0.816  &0.379  &0.175  &1.274\\
		 
		 Prompt 2  &0.377  &0.667  &0.845  &0.385  &0.161  &1.196\\
		 
		 Prompt 3 &0.380  &0.670  &0.846  &\textbf{0.375}  &0.160  &1.196\\

	\bottomrule
	\specialrule{0em}{7pt}{7pt}
	\end{tabular}
\end{adjustbox}
\caption{Ablations of prompt design. Perform in all classes of NYU Depth v2~\cite{silberman2012indoor}. The upper table shows partition detailed semantic token words of prompt design , the lower table shows the performance of each prompt design in all classes. The original prompt design is the one used for Table \ref{tab:nyu_results} to evaluate all classes. The best prompt design is bold.}
\vspace*{-3pt}
\label{tab:prompt_design}
\end{table*}

\subsection{Semantic Bin Responses}
\begin{figure}[t]
  \centering
\includegraphics[width=0.48\textwidth]{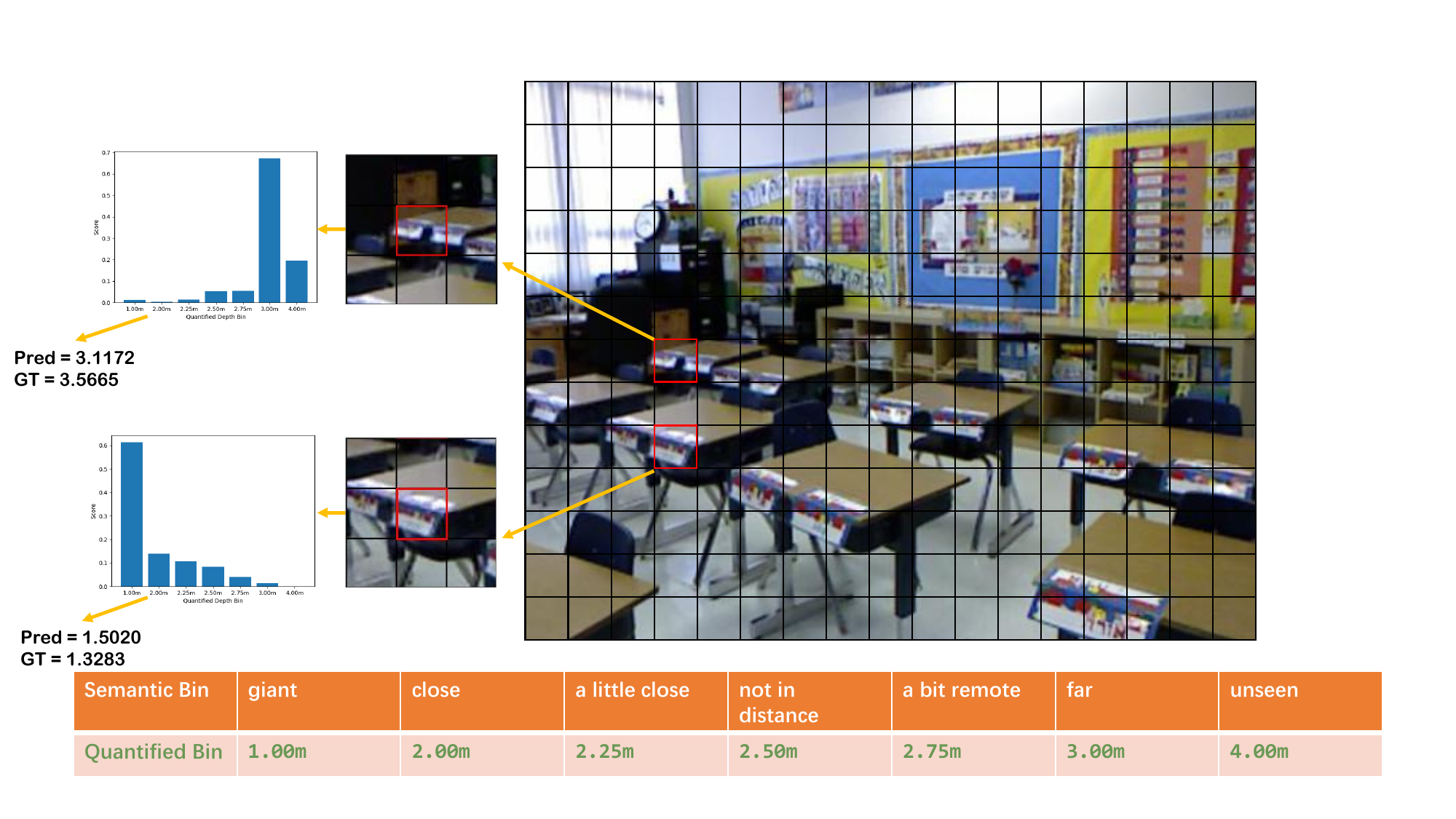}
  \caption{Semantic bin responses in different patches of an image for DepthCLIP.}
    \label{fig:vis}
    \vspace{-0.5cm}
\end{figure}

Shown as Figure\ref{fig:vis}, we visualize semantic bin responses in different patches within one single image, to demonstrate CLIP~\cite{clip}'s ability to distinguish patches from different distances. Here, the final feature map would be 13*17, then each patch of such feature map would be given a depth prediction. Each patch has a wide perceptive field during encoding, which enables it to catch surrounding information to decide its distance. Each patch would have its semantic bin responses then be attached to a certain depth and linearly combined to obtain final predication. Detailed correspondence between semantic bins and quantified depth is shown in the orange table. Each patch's response to the semantic bin is shown in the bar chart, and linearly combined predication with ground truth depth label is shown as "GT" and "Pred". We could notice a significant difference in semantic response for close patch and distant patch, which forms the foundation of the effectiveness of our DepthCLIP.

\subsection{Depth Predication Visualization}
\begin{figure}[t]
  \centering
\includegraphics[width=0.42\textwidth]{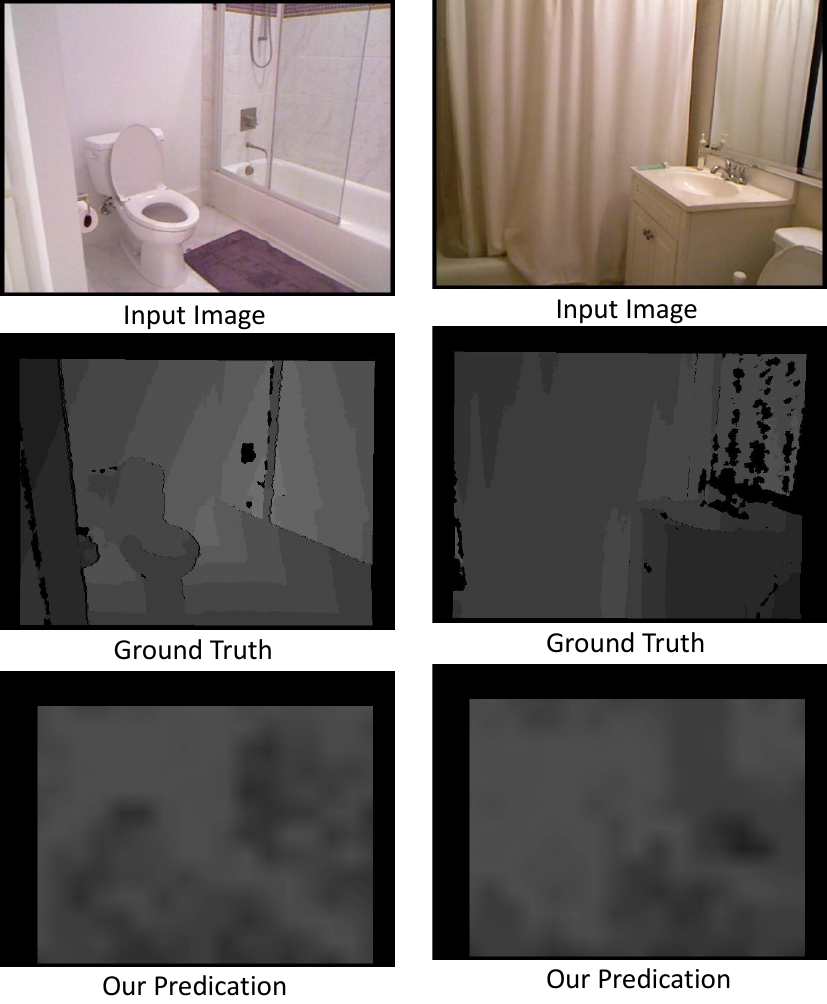}
  \caption{Depth predication visualization, compared with ground truth and input RGB image.}
    \label{fig:depth_pred}
\end{figure}

As shown in Figure \ref{fig:depth_pred}, we visualize depth predictions of our DepthCLIP, compared with ground truth and input RGB images. It could be witnessed that our prediction is a little bit blurred, and focuses more on detailed objects, like the tap on the left column, and the bathtub on the right column. This is discussed in the limitation section below, that the visual encoder of CLIP~\cite{clip} is pre-trained using a classification pre-text task, in which the model pays more attention to local details that help classification. In this way, backgrounds and local areas would be neglected.

\section{Limitations \& Future Research Direction}
\label{sec:limitations}
One limitation is that our work requires hand-craft depth bin tuning. For various depth distributions of different scenes, DepthCLIP would perform poorly if different scenes share the same depth bin setting. One possible solution is to conduct zero-shot image classification first, which could naturally be done by CLIP~\cite{clip}, then determine class-dependent depth bins for predicated scene class. Learning how to generate class-dependent depth bins for each predicated scene class would be a worthwhile future direction. 

Another limitation is its visual encoder of CLIP~\cite{clip} is trained under a classification-alike pre-training task. As a result, its visual encoder would pay more attention to local details which help identify its scene class. Local regions without noteworthy features for classification would produce non-significant semantic depth reactions, e.g., large areas of wall, ceiling, or floor. 
One possible solution is to train a vision language model with a regional pre-text task like image segmentation or object detection, to drive more regions' sensitivity.

\section{Conclusion}
\label{sec:conclusion}
In this paper, we propose DepthCLIP to conduct zero-shot monocular depth estimation with Contrastive Language-Image Pre-Training, which discards the need of training and data with depth labels. It directly applies CLIP~\cite{clip} that only pre-trained with large-scale image-text pairs to monocular depth estimation task, and achieved satisfying performance. We hope our work could cast a light on the research of bridging semantic vision language knowledge to quantitative vision tasks, and open future research on zero-shot monocular depth estimation. 

\paragraph{{\rm{\bf Acknowledgement.}}}
This work is supported by the National Natural Science Foundation of China (No.61971005), the R\&D Program of the Shaanxi Province of China (No.2022GY-064).

\bibliographystyle{ACM-Reference-Format}
\balance
\bibliography{sample-base}

\end{document}